\pdfoutput=1 

\documentclass[11pt]{article}

\usepackage[margin=1in]{geometry}
\usepackage{microtype}
\usepackage{times}            
\usepackage{helvet}\usepackage{courier}
\usepackage{setspace}
\usepackage[labelfont=bf]{caption}
\usepackage{enumitem}\setlist{nosep}

\usepackage{amsmath,amssymb,amsthm}
\usepackage{graphicx}
\usepackage{booktabs}
\usepackage{multirow}
\usepackage{siunitx}
\usepackage{wrapfig}
\usepackage{subcaption}
\usepackage{bigstrut}
\usepackage{multirow}
\usepackage{makecell}
\usepackage{algorithm}
\usepackage{algorithmic}
\usepackage[colorlinks=true, citecolor=blue, linkcolor=black, urlcolor=blue]{hyperref}
\usepackage[nameinlink]{cleveref}

\usepackage[sort&compress,round]{natbib}
\usepackage{authblk}
\title{MedSAMix: A Training-Free Model Merging Approach for Medical Image Segmentation}
\author[1,2]{Yanwu Yang\thanks{Equal contribution.}}
\author[3]{Guinan Su$^*$}
\author[4,5]{Jiesi Hu$^*$}
\author[1]{Francesco Sammarco} 
\author[3,6,7]{ \\Jonas Geiping\thanks{Senior authors}}
\author[1,2]{Thomas Wolfers$^\dag$}

\affil[1]{University of Tübingen, Germany}
\affil[2]{German Center for Mental Health (DZPG), partner site, Jena \& Tübingen, Germany}
\affil[3]{Max Planck Institute for Intelligent Systems, Germany} 
\affil[4,5]{Harbin Institute of Technology, Shenzhen, China; Peng Cheng Laboratory, Shenzhen, China}
\affil[6,7]{ELLIS Institute Tübingen, Germany; Tübingen AI Center, Germany}
\affil[ ]{\texttt{yangyanwu1111@gmail.com}}

\date{}  

\begin{document}
\maketitle
\begin{abstract}
Universal medical image segmentation models have emerged as a promising paradigm due to their strong generalizability across diverse tasks, showing great potential for a wide range of clinical applications. This potential has been partly driven by the success of general-purpose vision models such as the Segment Anything Model (SAM), which has inspired the development of various fine-tuned variants for medical segmentation tasks. However, fine-tuned variants like MedSAM are trained on comparatively limited medical imaging data that often suffers from heterogeneity, scarce annotations, and distributional shifts. These challenges limit their ability to generalize across a wide range of medical segmentation tasks.
In this regard, we propose MedSAMix, a training-free model merging method that integrates the strengths of both generalist models (e.g., SAM) and specialist models (e.g., MedSAM) for medical image segmentation.
In contrast to traditional model merging approaches that rely on manual configuration and often result in suboptimal outcomes, we propose a zero-order optimization method to automatically discover optimal layer-wise merging solutions.
Furthermore, for clinical applications, we develop two regimes to meet the demand of domain-specificity and generalizability in different scenarios by single-task optimization and multi-objective optimization respectively.
Extensive evaluations on 25 medical segmentation tasks demonstrate that MedSAMix effectively mitigates model bias and consistently improves performance in both domain-specific accuracy and generalization, achieving improvements of 6.67\% on specialized tasks and 4.37\% on multi-task evaluations.
\end{abstract}


\section{Introduction}
Universal medical image segmentation has become a widely adopted solution for diverse medical imaging tasks, enabling broad applicability without the need for extensive annotations or clinical expertise \cite{butoi2023universeg,czolbe2023neuralizer}. These generalizable models facilitate adaptability to domain shifts and even generalize to previously unseen tasks \cite{hu2025building}. 
Fine-tuned models derived from general-purpose foundation models for segmentation, such as the Segment Anything Model (SAM) \cite{kirillov2023segment}, represent a significant step toward universal medical image segmentation, e.g., MedSAM \cite{ma2024segment} and MedicoSAM \cite{archit2025medicosam}.

\begin{figure}[t]
\centering
\includegraphics[width=0.5\columnwidth]{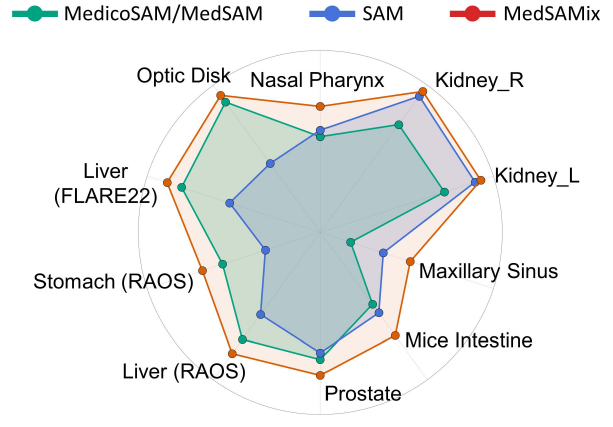} 
\caption{
Segmentation performance of various SAM-based models on diverse medical tasks, measured by Dice coefficient. Blue lines indicate the better result between MedSAM and MedicoSAM, while green lines represent SAM’s performance. Despite being domain-specific, MedSAM and MedicoSAM still underperform on certain tasks. In contrast, our MedSAMix (red) perform better consistently across tasks.
}
\label{fig1}
\end{figure}

Despite the promising results of these fine-tuned variants in universal medical image segmentation, we find their performance unbalanced across different tasks. As shown in Fig.~\ref{fig1}, MedSAM/MedicoSAM achieve strong results on certain familiar tasks, such as optic disk and liver segmentation. However, on others such as kidney and maxillary sinus segmentation they even underperform the original SAM (see additional results in the Results section). This highlights the stronger generalization capability of SAM, despite the absence of domain-specific optimization, and reveals the limited adaptability of its fine-tuned variants.

This can be mainly attributed to the inherent complexities of medical imaging data. Due to heterogeneity, domain shifts, and class imbalance, medical imaging datasets introduce complex optimization landscapes during fine-tuning, often leading models to converge to suboptimal local minima \cite{sanjeev2024fissionfusion,li2020domain}. These complexities make fine-tuned models such as MedSAM more susceptible to suboptimal generalization. In contrast, SAM, which is trained on large-scale natural image datasets with smoother optimization landscapes may retain stronger global generalization and thus outperform MedSAM on certain medical tasks. Furthermore, fine-tuned variants are often affected by catastrophic forgetting during adaptation, especially in the absence of strategies to preserve their original generalization capabilities. As a result, these models may lose part of their broader segmentation ability when adapted to the medical imaging domain \cite{kemker2018measuring,aleixo2023catastrophic}.

This raises a critical question: \textbf{How can we enhance domain-specific capabilities while mitigating the compromise of generalization?}
Noting that fine-tuned models initialized from the same pre-trained weights often converge to similar loss basins \cite{neyshabur2020being}, model merging has emerged as an effective strategy to unify diverse solution modes into a single model without additional training \cite{su2025fine,su2025gptailor,yadav2024ties}. By integrating parameters or representations from multiple models, model merging provides a promising approach to improve performance and mitigate single-model biases, resulting in more stable, diverse, and generalizable predictions \cite{almakky2024medmerge,akiba2025evolutionary}.
As medical models are often trained separately on data from different clinical centers due to privacy constraints, model merging offers a promising solution for effectively integrating these models without requiring data sharing, and its potential in the medical domain remains largely underexplored. Moreover, most existing model merging methods in other domains either rely on manually crafted configurations, which often result in suboptimal performance \cite{maron2022model, yadav2024ties}, or require computationally intensive merging during the training process \cite{sanjeev2024fissionfusion, qazi2024dynammo}, making them particularly costly for large foundation models such as SAM and MedSAM. In addition, these methods typically lack support for multi-objective optimization, making it difficult to ensure the generalization ability of the merged models.



In this study, we propose MedSAMix, an efficient training-free model merging framework to balance generalization and domain-specific capabilities for SAM-based medical image segmentation. Specifically, we explore the potential of MedSAMix from two perspectives:
(1) \textbf{Expert capability:} MedSAMix merges model variants through single-task optimization using only a few calibration samples, tailored to task-specific distributions and improved performance without retraining.
(2) \textbf{General capability:} We introduce a multi-objective optimization to capture diverse aspects of model performance across tasks for improving universal medical image segmentation.

MedSAMix employs a zero-order optimization approach that selects merging configurations based on their empirical performance during the search, enabling efficient exploration of the solution space given only a few samples. This allows MedSAMix to adaptively balance task-specificity and generalization by combining model variants, while mitigating suboptimal generalization and fine-tuning issues.
In summary, our main contributions are as follows:

\begin{itemize}
    \item We propose MedSAMix, a training-free model merging method leveraging the strengths of general and expert knowledge for medical image segmentation.

    \item We introduce flexible ways of merging for MedSAMix to accommodate different scenarios: single-task merging focuses on expert capabilities for specific domain tasks, while multi-task merging promotes generalization across diverse tasks by jointly optimizing multiple objectives.
    
    \item  Extensive experiments on 25 medical image segmentation tasks show that MedSAMix facilitates the enhancement of expert-level performance and generalization without retraining, achieving improvements of 6.67\% on specialized tasks and 4.37\% on multi-task evaluations.
    
\end{itemize}


\section{Related Work}
\subsection{Universal Medical image segmentation}
Recently, universal models for medical image segmentation have become powerful tools for medical image segmentation without retraining \cite{ma2024segment, butoi2023universeg,zhao2025segmic}. These models typically leverage image-mask pairs or symbolic prompts such as bounding boxes and points, offering strong zero-shot generalization capabilities. Among them, in-context learning (ICL) methods \cite{hu2024icl,takaya2024context,gao2025show} relate rich spatial alignment information by utilizing image-mask pairs as prompts, including UniverSeg \cite{butoi2023universeg}, SegGPT \cite{wang2023seggpt}, Neuralizer \cite{czolbe2023neuralizer}, Tyche \cite{rakic2024tyche} and Neuroverse3D \cite{hu2025building}. These models have achieved accurate segmentation performance on unseen data or tasks. However, these models are inherently constrained by their small parameter scale and the computational cost of in-context learning, both of which limit their generalizability.

In addition, building on the Segment Anything Model (SAM), variants including MedSAM \cite{ma2024segment}, SAM-Med \cite{ye2023sa}, MedicoSAM \cite{archit2025medicosam} have been proposed by fine-tuning on large-scale medical imaging datasets with more parameters, showing promising performance in various medical tasks. Compared to ICL-based models, SAM-based approaches benefit from larger model capacity and broader training coverage, which contribute to their superior generalization. However, the success of such foundation models often comes at the cost of large data requirements and substantial training overhead. In contrast, our proposed MedSAMix is an efficient and training-free approach that operates at the model level, reducing dependence on large-scale datasets while preserving strong generalization across diverse segmentation tasks.

\begin{figure}[t]
\centering
\includegraphics[width=1\columnwidth]{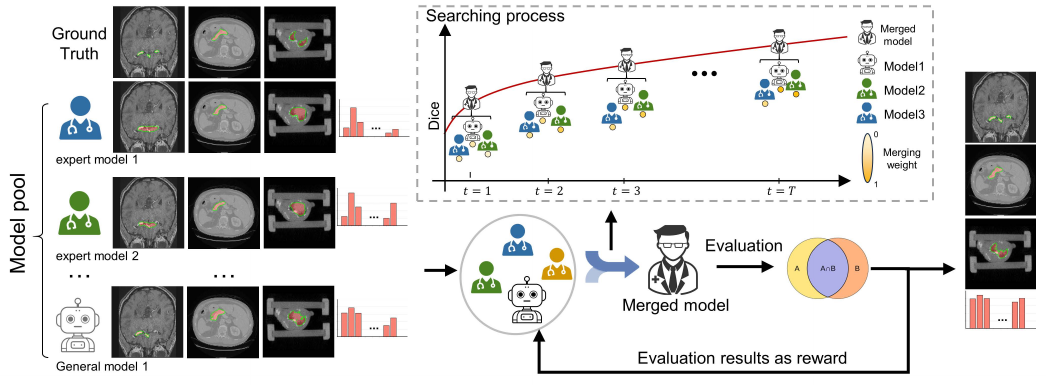} 
\caption{Overview of our model merging framework. Given a pool of models, MedSAMix searches for optimal merging configurations by using single-task or multi-task performance as rewards. While individual models may exhibit varying behaviors across tasks, MedSAMix adaptively combines them to optimize performance for the target task.
}
\label{fig2}
\end{figure}

\subsection{Model Merging} 
Early model merging relied on direct weight averaging \citep{utans1996weight, white2016sampling}, which lacked fine-grained control over model behaviors. Subsequent methods introduced parameter-space transformations, either through adjustment matrices \citep{matena2022merging, jin2022dataless} or task vectors defined by fine-tuning deltas \citep{ilharco2022editing}, enabling algebraic composition of capabilities across tasks. To address parameter conflicts, sparsity-driven strategies like TIES-Merging \citep{yadav2024ties} and DARE \citep{yu2024language} selectively retain and rescale parameters based on magnitude. More recent approaches explored parameter-level hyperparameter control \citep{yang2023adamerging, du2024parameter}, though still limited to task-vector formulations, while evolutionary merging \citep{akiba2024evolutionary} suffers from high search complexity. In this study, our framework integrates multiple merging approaches with fine-grained hyperparameter control tailored for image segmentation. Unlike prior medical imaging approaches \cite{qazi2024dynammo,sanjeev2024fissionfusion}, which merge networks during training to produce a robust model, our method adopts a fundamentally different paradigm: MedSAMix enables efficient post-hoc merging across different SAM fine-tuned models.

\section{Methodology}
In this section, we present a comprehensive overview of our approach. We start by formalizing the problem and introducing an optimization framework built upon three core components: (1) Search Space: We design a layer-wise search space specifically for SAM-series models based on Vision Transformers (ViT). This space allows for fine-grained control over merging strategies by supporting multiple merging methods at varying layer granularities across key modules, including the image encoder, prompt encoder, and mask decoder. (2) Optimization Objectives: The framework accommodates both task-specific objectives, enabling targeted enhancements for specialized domains, and multi-objective formulations that aim to identify Pareto-optimal configurations across diverse segmentation tasks. (3) Search Algorithm: We adopt SMAC optimization algorithm \cite{lindauer2022smac3}, which leverages the defined objectives as rewards to steer the search toward effective merge configurations. The following subsections elaborate on each component and explain how they collectively contribute to robust model merging for segmentation applications.

\subsection{Problem Formulation} 
Given a pre-trained segmentation base model $M_{\text{base}}$ and a set of candidate segmentation models $\mathcal{M} = \{M_1, M_2, ..., M_K\}$ fine-tuned from the same base architecture, our goal is to construct an optimal merged model that maximizes performance across a single dataset or multiple datasets.

These combinations are determined by a set of hyperparameters $\omega \in \Omega$, where $\Omega$ represents the search space of all possible merging configurations. Each configuration $\omega$ defines a specific strategy to combine components from candidate models to form a merged model $M_\omega$. The performance of the merged model is evaluated using an objective function $f(M_\omega)$ that measures effectiveness across segmentation tasks. This leads to our optimization problem:
\begin{equation}
\omega^* = \arg\min_{\omega \in \Omega} f(M_\omega)
\end{equation}
where $\omega^*$ represents the optimal hyperparameter configuration that yields the best-performing merged model according to the chosen objective.

\subsection{Search Space}
Our candidate models are all based on the Segment Anything Model (SAM) architecture \cite{kirillov2023segment}. The network is built on the transformer architecture, specifically incorporating a Vision Transformer-based \cite{dosovitskiy2020image} image encoder with $l$ transformer layers responsible for extracting image features, a prompt encoder with $k$ convolutional downsampling layers for integrating user interactions, and a lightweight mask decoder with $z$ transformer layers and $p$ transposed convolutional layers that generate segmentation results and confidence scores. The search space $\Omega$ encompasses all possible configurations for constructing our merged model. For different components (image encoder, prompt encoder, and mask decoder), we employ varying layer granularity to determine how layers are grouped for merging, where each group shares the same merge hyperparameters. This granularity-based approach allows us to balance the size of the search space with fine-grained control over the merging process. Specifically, we define granularity $g_{\text{enc}}$, $g_{\text{prompt}}$, and $g_{\text{dec}}$ for the three components respectively, dividing the layers into $G = \lceil l/g_{\text{enc}} \rceil + \lceil k/g_{\text{prompt}} \rceil + \lceil (z+p)/g_{\text{dec}} \rceil$ groups in total. For each layer group $i$, we specify a merge method $d_i \in \{1, 2, \ldots, D\}$ selected from $D$ available merging techniques, and associated hyperparameters $\mathbf{h}_i = [h_{i,1}, h_{i,2}, \ldots, h_{i,P_i}]$, where $P_i$ is the number of hyperparameters for merge method $d_i$. Therefore, a complete configuration $\omega \in \Omega$ is represented as:
\begin{equation}
    \omega = \{\{(d_1, \mathbf{h}_1), (d_2, \mathbf{h}_2), \ldots, (d_G, \mathbf{h}_G)\}\}
\end{equation}

This formulation provides a structured framework for exploring the space of possible model merging configurations while maintaining computational tractability through hierarchical grouping.

\subsection{Optimization Objective}
To evaluate the quality of the merged model, we define both single-objective and multi-objective optimization that measures the model's effectiveness across tasks. Specifically, we measure performance on calibration datasets $\mathcal{D}$, quantifying metrics such as Dice coefficient for segmentation.

\textbf{For single-task optimization,} we focus on maximizing performance on a specific calibration target task $T$:
\begin{equation}
f_{\text{single} }(M_\omega) = \mathcal{L}(M_\omega, \mathcal{D}_T)
\end{equation}
where $\mathcal{L}$ represents the segmentation loss function (e.g., Dice loss, cross-entropy) evaluated on the target dataset $\mathcal{D}_T$.

\textbf{For multi-task optimization} across tasks $\mathcal{T} = \{T_1, T_2, \ldots, T_m\}$, we employ Pareto Efficient Global Optimization (ParEGO) \citep{knowles2006parego} to identify Pareto-optimal solutions:
\begin{equation}
\begin{aligned}
    f_{\text{multi} }(M_\omega, \lambda) = & \max_{i=1,\ldots,m} \{\lambda_i \cdot f_{\text{single},i}(M_\omega)\} \\ &+ \alpha \sum_{i=1}^{m} \lambda_i \cdot f_{\text{single},i}(M_\omega)
\end{aligned}
\end{equation}
where $f_{\text{single},i}(M_\omega)$ is the $i$-th objective function, $\lambda_i$ is the corresponding weight satisfying $\sum_{i=1}^{m} \lambda_i = 1$ and $\lambda_i \geq 0$, and $\alpha$ is a small positive constant (typically 0.05). Each task-specific objective is defined as:
\begin{equation}
f_{\text{single},i}(M_\omega) = \mathcal{L}_i(M_\omega, \mathcal{D}_{T_i})
\end{equation}
The optimizer outputs a Pareto front of merging configurations representing different trade-offs between tasks. For evaluation, we selected the configurations yielding the best Pareto front on the calibration set.

\subsection{Search Algorithm}
To efficiently navigate the large search space $\Omega$ and find optimal merging configurations, we employ Bayesian optimization based on SMAC \cite{lindauer2022smac3} with Random Forest \cite{breiman2001random} as the surrogate model. Given evaluated configurations $\mathcal{H}_t = \{(\omega_1, f(M_{\omega_1})), \ldots, (\omega_t, f(M_{\omega_t}))\}$ at iteration $t$, the next configuration is selected by:

\begin{equation}
\omega_{t+1} = \arg\max_{\omega \in \Omega} \text{EI}(\omega)
\end{equation}

where ${\text{EI}}(\omega) = \mathbb{E}[\max(f^* - \hat{f}(\omega), 0)]$ is the Expected Improvement acquisition function, $f^*$ is the best observed value, and $\hat{f}(\omega)$ is the Random Forest prediction. The process iteratively selects promising configurations until convergence. The whole process is described in Alg. \ref{alg:model_merging}

\begin{algorithm}[h]
\caption{Optimization Process of MedSAMix}
\label{alg:model_merging}
\begin{algorithmic}[1]
\REQUIRE Base model $M_{\text{base}}$, candidate models $\mathcal{M} = \{M_1, M_2, \ldots, M_K\}$, calibration datasets $\mathcal{D}$, maximum iterations $T_{\max}$
\ENSURE Optimal merging configuration $\omega^*$

\STATE Initialize search space $\Omega$ with granularities $g_{\text{enc}}, g_{\text{prompt}}, g_{\text{dec}}$
\STATE Initialize evaluation history $\mathcal{H}_0 = \emptyset$
\STATE Randomly sample initial configurations and evaluate to get $\mathcal{H}_1$

\FOR{$t = 1$ to $T_{\max}$}
    \STATE Train Random Forest surrogate model on $\mathcal{H}_t$
    \STATE Compute Expected Improvement: ${\text{EI}}(\omega) = \mathbb{E}[\max(f^* - \hat{f}(\omega), 0)]$
    \STATE Select next configuration: $\omega_{t+1} = \arg\max_{\omega \in \Omega} {\text{EI}}(\omega)$
    
    \STATE \textbf{// Model Merging based on Configuration $\omega_{t+1}$}
    \FOR{each layer group $i = 1$ to $G$}
        \STATE Apply merge method $d_i$ with hyperparameters $\mathbf{h}_i$
        \STATE Merge corresponding layers from candidate models
    \ENDFOR
    
    \STATE \textbf{// Evaluation}
    \IF{single-task optimization}
        \STATE Compute $f_{\text{single}}(M_{\omega_{t+1}}) = \mathcal{L}(M_{\omega_{t+1}}, \mathcal{D}_T)$
    \ELSE
        \STATE Compute $f_{\text{multi}}(M_{\omega_{t+1}}, \lambda)$ using ParEGO on tasks $\mathcal{T} = \{T_1, T_2, \ldots, T_m\}$
    \ENDIF
    
    \STATE Update history: $\mathcal{H}_{t+1} = \mathcal{H}_t \cup \{(\omega_{t+1}, f(M_{\omega_{t+1}}))\}$
    \STATE Update best configuration: $f^* = \min_{\omega \in \mathcal{H}_{t+1}} f(M_\omega)$
\ENDFOR

\RETURN $\omega^* = \arg\min_{\omega \in \mathcal{H}_{T_{\max}}} f(M_\omega)$
\end{algorithmic}
\end{algorithm}

\begin{table}[ht]
  \centering
  \fontsize{9pt}{9.5pt}\selectfont
  \caption{Segmentation performance across 25 tasks in terms of Dice coefficient score (\%). The values in parentheses in the last row indicate the number of tasks in which each method outperforms the baseline MedicoSAM/MedSAM/SAM model.}

  \setlength{\tabcolsep}{1.5pt}
\begin{tabular}{cccccccccccccccc}
\hline
\hline
\multirow{2}[3]{*}{idx} &       & \multirow{2}[3]{*}{Task Name} &       & Supervised &       & \multicolumn{3}{c}{In-context-learning models} &       &       & \multicolumn{5}{c}{SAM-based} \bigstrut[b]\\
\cline{5-5}\cline{7-9}\cline{11-16}      &       &       &       & nnU-Net &       & SegGPT & UniverSeg & Neuro3D &       & SAM-Med & \makecell{Medico \\ SAM} & MedSAM & SAM   & Ours-S & Ours-M \bigstrut\\
\cline{1-1}\cline{3-3}\cline{5-5}\cline{7-9}\cline{11-16}1     &       & Brain Tumor &       & 75.83 &       & 17.24 & 19.20 & 68.05 &       & 67.26 & 73.17 & 70.56 & 70.04 & \textbf{78.36} & \underline{75.18} \bigstrut[t]\\
2     &       & Vascular &       & 85.87 &       & 28.33 & \underline{68.21} & \textbf{80.27} &       & 36.85 & 31.35 & 28.80 & 50.86 & 60.70 & 62.90 \\
3     &       & Cerebral Cortex &       & 87.72 &       & 47.08 & \underline{69.51} & \textbf{87.08} &       & 31.77 & 56.13 & 48.83 & 55.00 & 58.00 & 55.03 \\
4     &       & Hippocampus &       & 82.30 &       & 25.52 & \underline{71.33} & \textbf{75.92} &       & 36.17 & 52.17 & 43.36 & 53.23 & 62.14 & 57.51 \\
5     &       & Thalamus &       & 83.40 &       & 38.70 & \underline{74.57} & \textbf{78.09} &       & 56.49 & 67.60 & 56.99 & 44.08 & 70.98 & 67.60 \\
6     &       & Lateral Ventricle &       & 84.76 &       & 45.60 & 75.23 & \textbf{82.44} &       & 53.83 & 55.74 & 27.52 & 74.68 & 75.71 & \underline{77.60} \\
7     &       & Putamen &       & 83.03 &       & 21.92 & \underline{73.38} & \textbf{78.41} &       & 25.74 & 36.93 & 26.26 & 32.49 & 45.33 & 33.61 \\
8     &       & Amygdala &       & 73.43 &       & 7.77  & \underline{59.37} & 57.85 &       & 30.33 & 48.62 & 39.58 & 29.18 & \textbf{61.92} & 52.03 \\
9     &       & FLARE22 Liver &       & 96.59 &       & 68.72 & 79.33 & 90.73 &       & 92.03 & 91.83 & 85.89 & 84.60 & \textbf{94.05} & \underline{92.88} \\
10    &       & FLARE22 Kidney\_R &       & 94.66 &       & 63.24 & 84.21 & 76.34 &       & 93.95 & 90.04 & 84.95 & 95.05 & \textbf{95.86} & \underline{95.12} \\
11    &       & FLARE22 Kidney\_L &       & 95.39 &       & 66.75 & 83.33 & 79.25 &       & 92.92 & 89.64 & 83.87 & \underline{94.25} & \textbf{95.06} & 94.16 \\
12    &       & Maxillary Sinus &       & 91.23 &       & 51.30 & \underline{80.57} & 60.29 &       & 60.90 & 75.54 & 56.40 & 80.45 & \textbf{84.47} & 79.61 \\
13    &       & Nasal Cavity &       & 88.37 &       & 39.83 & \textbf{69.76} & 60.36 &       & 54.23 & 60.15 & 40.42 & 58.70 & 62.51 & \underline{65.29} \\
14    &       & Nasal Pharynx &       & 91.40 &       & 46.35 & 84.87 & 83.61 &       & 87.44 & 84.68 & 70.52 & 85.62 & \textbf{89.01} & \underline{88.83} \\
15    &       & Prostate &       & 86.78 &       & 50.26 & 76.36 & 33.57 &       & 88.40 & 89.16 & 79.01 & 88.21 & \textbf{91.36} & \underline{90.64} \\
16    &       & Mice-Lung &       & 89.28 &       & 62.17 & 75.67 & \textbf{83.25} &       & 61.08 & 65.93 & 49.04 & 71.96 & \underline{80.62} & 75.43 \\
17    &       & Mice-Pancreas &       & 83.27 &       & 53.52 & 65.21 & 66.03 &       & 85.35 & 83.68 & 74.17 & 85.20 & \textbf{89.20} & \underline{88.44} \\
18    &       & Cardiac &       & 85.00 &       & 43.23 & 68.96 & 64.21 &       & 77.06 & 74.31 & 66.18 & 74.87 & \textbf{84.41} & \underline{83.03} \\
19    &       & FLARE22 Spleen &       & 95.14 &       & 64.57 & 70.59 & 82.61 &       & 93.46 & 92.93 & 83.86 & 92.60 & \textbf{95.06} & \underline{94.76} \\
20    &       & FLARE22 Pancreas &       & 79.92 &       & 2.36  & 36.38 & 16.89 &       & 71.66 & 76.59 & 58.14 & 70.26 & \underline{77.66} & \textbf{78.36} \\
21    &       & RAOS Liver &       & 93.74 &       & 70.81 & 81.49 & 81.83 &       & 88.19 & \underline{89.90} & 84.64 & 85.50 & \textbf{92.38} & 89.82 \\
22    &       & RAOS Kidney &       & 92.39 &       & 63.14 & \textbf{80.14} & 39.74 &       & 62.34 & 72.44 & 66.01 & 69.13 & \underline{74.66} & 72.72 \\
23    &       & RAOS Stomach &       & 87.73 &       & 33.03 & 57.92 & 33.50 &       & 79.85 & 85.67 & 77.64 & 79.23 & \textbf{88.74} & \underline{86.00} \\
24    &       & Optic Cup &       & 85.68 &       & 74.32 & 80.56 & -     &       & 71.61 & \underline{86.42} & 86.41 & 61.93 & \textbf{87.70} & 82.11 \\
25    &       & Optic Disk &       & 96.67 &       & \textbf{96.27} & \underline{95.61} & -     &       & 85.53 & 93.70 & 94.01 & 83.17 & 95.17 & 94.73 \bigstrut[b]\\
\hline
      &       & Avg.  &       & 87.58 &       & 47.28 & 71.27 & 67.84 &       & 67.38 & 72.97 & 63.32 & 70.81 & 79.64 (25) & 77.34 (18) \bigstrut\\
\hline
\hline
\end{tabular}%

\begin{flushleft}
 \small \textit{*Ours-S:} MedSAMix-S. The model is merged under a specific single-task setting and evaluated on the testing set. \\
\textit{*Ours-M:} MedSMix-M. The model is merged under a multi-task setting and evaluated on the test data across all 25 tasks.
\end{flushleft}
  \label{main_res}%
\end{table}%

\section{Experiments}

\textbf{Datasets.}
In this study, we incorporate 25 publicly available medical image segmentation datasets across a wide range of organs (e.g., liver, kidney, spleen) and imaging modalities (e.g., CT, MRI, MRA).
These datasets cover diverse tasks, including brain tumor segmentation (BraTS) \cite{menze2014multimodal}, vascular segmentation (Topcow) \cite{topcowdata}, optic disc and cup segmentation from retinal fundus images (Fundus) \cite{staal2004ridge}, and abdominal organ segmentation for the kidneys, liver, spleen, pancreas (FLARE22) \cite{ma2024unleashing}, liver, kidney, and stomach (RAOS) \cite{luo2024rethinking}. We also include maxillary sinus, nasal cavity segmentation, and nasal pharynx (Nasal) \cite{zhang2024nasalseg}, cardiac structure segmentation (MSD) \cite{antonelli2022medical}, and prostate segmentation (PROMISE) \cite{litjens2014evaluation}.
For a broad and fair comparison as well as to enhance the diversity, we include additional commonly studied segmentation tasks, such as body organ segmentation in rodents (e.g., lungs and intestines) using the Mice dataset \cite{rosenhain2018preclinical}, as well as various neuroanatomical structure segmentation from the ADNI dataset \cite{jack2008alzheimer}, including the cerebral cortex, hippocampus, thalamus, amygdala, lateral ventricles, and putamen. More details of the data size are reported in the Supplementary.

\textbf{Benchmarks.}
We compare our method with ICL-based baselines, including UniverSeg \cite{butoi2023universeg}, SegGPT \cite{wang2023seggpt}, and Neuroverse3D (Neuro3D) \cite{hu2025building}, as well as SAM-based models such as SAM-Med2D \cite{ye2023sa}, SAM \cite{kirillov2023segment}, MedSAM \cite{ma2024segment}, and MedicoSAM \cite{archit2025medicosam}. In addition, we use the fully-supervised nnU-Net \cite{isensee2021nnu} as an upper bound for reference. Additional comparisons are supplied in the Supplementary.

\textbf{Evaluations.}
For each dataset, 80\% is reserved for testing to ensure robust evaluation. The remaining 20\% serves multiple purposes: as a calibration set for MedSAMix during the merging search, as context prompts for ICL-based models, and as training data for nnU-Net. SAM-based models are evaluated in a zero-shot setting without access to training data but are provided with inferred bounding boxes for guidance. All tasks are evaluated using the Dice coefficient.



\textbf{Model Selection.}
We use SAM \cite{kirillov2023segment} as the base and merge it with MedSAM \cite{ma2024segment} and MedicoSAM \cite{archit2025medicosam} as representative fine-tuned variants.
Although other variants exist, such as the Medical SAM Adapter \cite{wu2025medical}, they differ architecturally from SAM, making them incompatible with our model merging framework.


\textbf{Objective and Optimizer.}
For expert capability evaluation, we focus on single-task search. For each task, the model is optimized using the calibration set to obtain the optimal merging configuration, which is then evaluated on the corresponding test set. This procedure is repeated for all 25 tasks.
For general capability evaluation, we perform multi-objective optimization across a representative subset of tasks to reflect overall performance. To ensure both diversity and representativeness, we select eight tasks for the search process: optic disk, tumor, vascular, lateral ventricle, mice-lung, cardiac, nasal pharynx, and prostate. The merged models obtained from single-task and multi-task settings are referred to as \textbf{MedSAMix-S} and \textbf{MedSAMix-M}, respectively.

\begin{figure}[h]
\centering
\includegraphics[width=1\columnwidth]{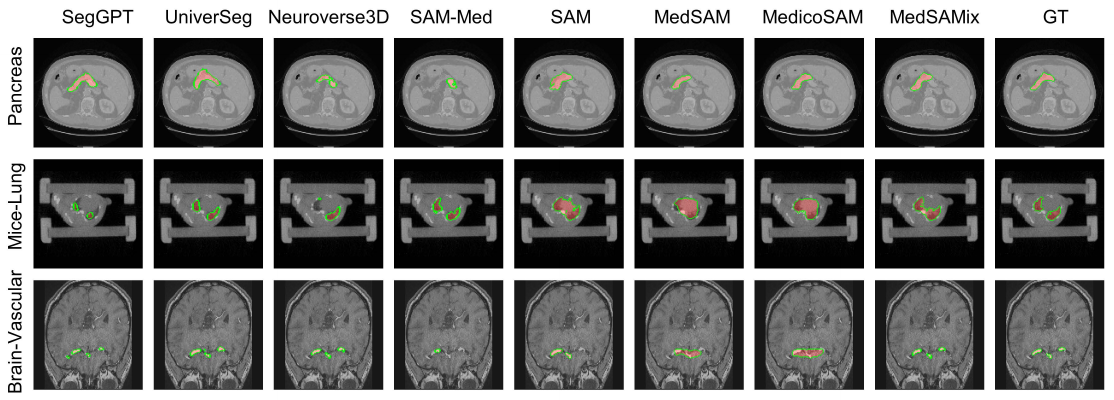} 
\caption{Visual examples of segmentation results of implemented models on pancreas, mouse lung, and brain vascular datasets.}
\label{fig4}
\end{figure}

\textbf{Implementation details.}
In this study, we leverage the commonly used base architectures of SAM and its variants, MedSAM, and MedicoSAM for model merging, with $l=12$ Transformer layers, 768 hidden features, 12 heads, $k=4$ prompt encoder layers, and a lightweight mask decoder with $z=2$ Transformer layers and $p=3$ transposed convolutional layers. Our model merging framework adaptively searches for optimal layer combinations, with candidate merging techniques including TIES, task arithmetic, linear combination, and SLERP. For single-task optimization, we perform 120 trials using two GPUs, while for multi-task optimization, we conduct 200 trials on four GPUs. The layer granularity is searched within the range [1, 4]. Codes and model weights would be available upon acceptance. Our code and checkpoints are available at Github \footnote{\url{https://github.com/podismine/MedSAMix.git}} and Hugging Face\footnote{\url{https://huggingface.co/guinansu/MedSAMix}}.



\section{Results}

\subsection{Evaluations on domain-specific single tasks}
\label{sec_res1}

We first evaluated the performance in domain-specific single-task applications. The results are shown in Table~\ref{main_res}, where our merged model is denoted as MedSAMix-S. From the results, we can see that MedSAMix-S achieves significant improvements over the original SAM and other fine-tuned models across all the tasks. For instance, on the brain tumor and Nasal Pharynx parcellation tasks, MedSAMix-S outperforms the MedicoSAM by 5.19\% and 4.33\% in terms of Dice coefficient, respectively.
Moreover, for tasks where MedSAM and MedicoSAM already perform well, such as optic disk segmentation, MedSAMix-S still yields further improvements of around 1.2\%. This demonstrates the advantage of model merging in leveraging the complementary strengths of different models, even when those models are already well-adapted to the task.
In addition, we report the fully supervised results from nnU-Net \cite{isensee2021nnu} as an upper bound for reference. Compared to nnU-Net, although there remains a performance gap of around 8\%, our model achieves comparable or even better results on tasks such as brain tumor and prostate segmentation. This underscores the strong potential of foundation models, highlighting their promising few-shot capabilities.


\subsection{Evaluations on universal segmentation tasks}\label{sec_res2}

In addition to the single-task evaluations, we also compare merged MedSAMix-M with other ICL-based and SAM-based models across multiple tasks to assess its generalization capability in medical image segmentation. The results are reported as MedSAMix-M in Table~\ref{main_res}, where it outperforms baselines on 18 out of 25 tasks. Among all universal segmentation models, MedSAMix-M achieves the best overall performance, with an average improvement of 4.37\% over the second-best model. 
Furthermore, most ICL-based models perform even worse than the original SAM. As shown in Fig.~\ref{fig4}, visual comparisons further illustrate this trend. For example, while SegGPT performs well on certain tasks, its average performance across all tasks remains relatively low, suggesting that these domain-specific foundation models still exhibit limited generalization. These findings underscore the critical importance of generalization capability in medical image segmentation.

\begin{table}[h]
  \centering
  \setlength{\tabcolsep}{20pt}
  \renewcommand{\arraystretch}{0.9}
  \small
  \caption{Comparisons with other model merging methods on medical image segmentation in terms of the averaged Dice coefficient (\%) score across 25 tasks. }
\begin{tabular}{c|ccc}
\hline
\hline
Type  & Models & Avg.  & Improv. \bigstrut\\
\hline
\multirow{3}[2]{*}{Baseline} & MedSAM & 63.32 & - \bigstrut[t]\\
      & SAM   & 70.81 & - \\
      & MedicoSAM & 72.97 & - \bigstrut[b]\\
\hline
\multirow{6}[2]{*}{Merged} & TIES  & 66.31 & -6.66 \% \bigstrut[t]\\
      & TA  & 72.19 & -0.78 \% \\
      & Linear & 74.00 & 1.03 \% \\
      & SLERP & 75.28 & 2.31 \% \\
      & MedSAMix-S & 79.64 & 6.67 \% \\
      & MedSAMix-M & 77.34 & 4.37 \%\\
\hline
\hline
\end{tabular}%
  \label{res2}%
\end{table}%

\subsection{Comparisons with other model merging methods}

Given the effectiveness of model merging in medical image segmentation, we compare our MedSAMix with several model merging baselines, including Task-Arithmetic (TA) \cite{ilharco2022editing}, TIES \cite{yadav2024ties}, SLERP \cite{white2016sampling}, and linear weighted combination. which we used in our search space. The results are summarized in Table~\ref{res2}, which reports the average Dice coefficient across all 25 tasks, along with the corresponding improvements over the MedicoSAM baseline. Detailed results are provided in the Supplementary Material.
The results show that most model merging baselines offer limited improvement over MedicoSAM. In contrast, MedSAMix(-S/-M) consistently outperforms them. These findings highlight the effectiveness and adaptability of MedSAMix in identifying optimal merging configurations for medical image segmentation.


\subsection{Search space sensitivity analysis}

\begin{figure}[h]
\centering
\includegraphics[width=1\columnwidth]{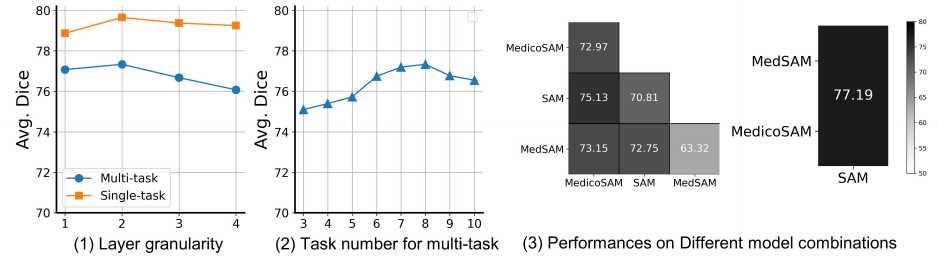} 
\caption{Sensitivity analysis of different merging settings in terms of (1) layer granularity, (2) task number for multi-task optimization, and (3) different model combinations.}
\label{fig3}
\end{figure}

\textbf{Layer granularity.} We conducted a sensitivity analysis on layer granularity for both single-task and multi-task optimization, varying the granularity from 1 to 4.
The results are presented in Fig.~\ref{fig3} (1), where the performance under single-task and multi-task settings is shown in orange and blue, respectively.
The results reveal that single-task performance is relatively insensitive to changes in layer granularity, whereas multi-task performance is notably affected. This suggests that for single-task scenarios, there remains redundancy in the network parameters, allowing for merging even with coarse granularity. In contrast, multi-task performance is more dependent on fine-grained control over layer merging. Notably, the best performance in the multi-task setting is achieved when the granularity is set to 2.

\noindent \textbf{Number of searched tasks.} In addition, we investigated the impact of the number of tasks used for multi-task optimization. As shown in Fig.~\ref{fig3} (2), we find a notable improvement when the number of search tasks exceeds six. The best performance is achieved when eight tasks are used. However, further increasing the number of tasks leads to a decline in performance. This may be attributed to the increasing difficulty in maintaining Pareto-optimal solutions and effective model selection as the task number grows.

\noindent \textbf{Combinations of models}. Finally, we compare model merging on different combinations of models, as shown in Fig.~\ref{fig3} (3). We find that merging MedicoSAM and SAM yields the best pairwise performance, consistent with their strong individual results.
Although MedSAM achieves the worst performance among the three models, combining all three models can lead to further improvements, demonstrating the benefit of leveraging complementary knowledge from multiple sources.
This demonstrates the great potential of our framework. Incorporating more diverse models is expected to further enhance performance.

\section{Discussion and Conclusion}
In this study, we propose MedSAMix, a zero-order training-free model merging method for medical image segmentation. While domain-specific foundation models achieve strong performance on specialized tasks, they often struggle with generalizability and, in some cases, underperform the original SAM model. In contrast, MedSAMix consistently outperforms all baselines across 25 tasks, demonstrating both superior domain-specific accuracy and broader generalization capabilities.

\noindent \textbf{Model merging for medical image analysis.} Our work represents a pioneering exploration of model merging in medical image analysis. In practice, individual centers can train models for specific organs or modalities and merge them with large-scale foundation models like SAM, yielding models that combine strong generalization with domain-specific expertise. Moreover, such a merging strategy offers a practical solution to challenges related to data privacy and security, as it eliminates the need for data sharing while integrating knowledge from diverse sources.

\noindent \textbf{Generalized vs. specialized models.} This study highlights the inherent trade-off between specialization and generalization in medical image segmentation. While expert models like MedicoSAM perform well across tasks, they still underperform the base model in certain cases. In contrast, the original SAM demonstrates strong generalizability, underscoring the value of foundation models. Based on this insight, our MedSAMix effectively balances this trade-off and enhances both multi-task generalization and single-task performance without the need for additional data or retraining. This highlights the power of model merging in mitigating suboptimal generalization and points to the potential of hybrid strategies that achieve performance gains purely through model-level optimization.

\noindent \textbf{Efficiency.}
Unlike fine-tuning methods that require memory-intensive training, MedSAMix merges models by directly searching optimal configurations without any gradient updates. Each merging configuration involves only forward inference, making the process highly parallelizable and GPU-efficient. 
For example, on the vascular segmentation task, MedSAMix completes 120 trials in 70 minutes using two GPUs, each consuming only 8GB of memory. This averages to 1.5 minutes per trial per GPU.
In the multi-task setting, 200 trials are executed over 20 hours on four GPUs.
In contrast, other universal segmentation models demand extensive resources and take days of training on 8 A100 GPUs (e.g., MedicoSAM and Neuroverse3D). Our MedSAMix, by comparison, is far more efficient, requiring minimal hardware and no retraining.

\noindent \textbf{Limitation.} We acknowledge that, despite covering 25 diverse tasks, our study cannot fully represent the entire landscape of medical image segmentation. Future work will expand task diversity and develop systematic methods to assess generalization and robustness.

\section{Acknowledgment}
This work was supported by the German Research Foundation (DFG) Emmy Noether Program (513851350, TW) and the BMBF/DLR project FEDORA (01EQ2403G, TW). We acknowledge the affiliation with the International Max Planck Research School for Intelligent Systems (IMPRS-IS) and the computational resources provided by the de.NBI Cloud, part of the German Network for Bioinformatics Infrastructure (de.NBI).

\begingroup
\small
\bibliography{aaai2026}  

\begin{thebibliography}{51}
\providecommand{\natexlab}[1]{#1}
\providecommand{\url}[1]{\texttt{#1}}
\expandafter\ifx\csname urlstyle\endcsname\relax
  \providecommand{\doi}[1]{doi: #1}\else
  \providecommand{\doi}{doi: \begingroup \urlstyle{rm}\Url}\fi

\bibitem[Akiba et~al.(2024)Akiba, Shing, Tang, Sun, and Ha]{akiba2024evolutionary}
Takuya Akiba, Makoto Shing, Yujin Tang, Qi~Sun, and David Ha.
\newblock Evolutionary optimization of model merging recipes.
\newblock \emph{arXiv preprint arXiv:2403.13187}, 2024.

\bibitem[Akiba et~al.(2025)Akiba, Shing, Tang, Sun, and Ha]{akiba2025evolutionary}
Takuya Akiba, Makoto Shing, Yujin Tang, Qi~Sun, and David Ha.
\newblock Evolutionary optimization of model merging recipes.
\newblock \emph{Nature Machine Intelligence}, 7\penalty0 (2):\penalty0 195--204, 2025.

\bibitem[Aleixo et~al.(2023)Aleixo, Colonna, Cristo, and Fernandes]{aleixo2023catastrophic}
Everton~L Aleixo, Juan~G Colonna, Marco Cristo, and Everlandio Fernandes.
\newblock Catastrophic forgetting in deep learning: A comprehensive taxonomy.
\newblock \emph{arXiv preprint arXiv:2312.10549}, 2023.

\bibitem[Almakky et~al.(2024)Almakky, Sanjeev, Hashmi, Qazi, Wang, and Yaqub]{almakky2024medmerge}
Ibrahim Almakky, Santosh Sanjeev, Anees Ur~Rehman Hashmi, Mohammad~Areeb Qazi, Hu~Wang, and Mohammad Yaqub.
\newblock Medmerge: merging models for effective transfer learning to medical imaging tasks.
\newblock \emph{arXiv preprint arXiv:2403.11646}, 2024.

\bibitem[Antonelli et~al.(2022)Antonelli, Reinke, Bakas, Farahani, Kopp-Schneider, Landman, Litjens, Menze, Ronneberger, Summers, et~al.]{antonelli2022medical}
Michela Antonelli, Annika Reinke, Spyridon Bakas, Keyvan Farahani, Annette Kopp-Schneider, Bennett~A Landman, Geert Litjens, Bjoern Menze, Olaf Ronneberger, Ronald~M Summers, et~al.
\newblock The medical segmentation decathlon.
\newblock \emph{Nature communications}, 13\penalty0 (1):\penalty0 4128, 2022.

\bibitem[Archit et~al.(2025)Archit, Freckmann, and Pape]{archit2025medicosam}
Anwai Archit, Luca Freckmann, and Constantin Pape.
\newblock Medicosam: Towards foundation models for medical image segmentation.
\newblock \emph{arXiv preprint arXiv:2501.11734}, 2025.

\bibitem[Breiman(2001)]{breiman2001random}
Leo Breiman.
\newblock Random forests.
\newblock \emph{Machine learning}, 45:\penalty0 5--32, 2001.

\bibitem[Butoi et~al.(2023)Butoi, Ortiz, Ma, Sabuncu, Guttag, and Dalca]{butoi2023universeg}
Victor~Ion Butoi, Jose Javier~Gonzalez Ortiz, Tianyu Ma, Mert~R Sabuncu, John Guttag, and Adrian~V Dalca.
\newblock Universeg: Universal medical image segmentation.
\newblock In \emph{Proceedings of the IEEE/CVF International Conference on Computer Vision}, pages 21438--21451, 2023.

\bibitem[Czolbe and Dalca(2023)]{czolbe2023neuralizer}
Steffen Czolbe and Adrian~V Dalca.
\newblock Neuralizer: General neuroimage analysis without re-training.
\newblock In \emph{Proceedings of the IEEE/CVF conference on computer vision and pattern recognition}, pages 6217--6230, 2023.

\bibitem[Dosovitskiy et~al.(2020)Dosovitskiy, Beyer, Kolesnikov, Weissenborn, Zhai, Unterthiner, Dehghani, Minderer, Heigold, Gelly, et~al.]{dosovitskiy2020image}
Alexey Dosovitskiy, Lucas Beyer, Alexander Kolesnikov, Dirk Weissenborn, Xiaohua Zhai, Thomas Unterthiner, Mostafa Dehghani, Matthias Minderer, Georg Heigold, Sylvain Gelly, et~al.
\newblock An image is worth 16x16 words: Transformers for image recognition at scale.
\newblock \emph{arXiv preprint arXiv:2010.11929}, 2020.

\bibitem[Du et~al.(2024)Du, Lee, Li, Jiang, Guo, Yu, Liu, Goh, Tang, He, et~al.]{du2024parameter}
Guodong Du, Junlin Lee, Jing Li, Runhua Jiang, Yifei Guo, Shuyang Yu, Hanting Liu, Sim~K Goh, Ho-Kin Tang, Daojing He, et~al.
\newblock Parameter competition balancing for model merging.
\newblock \emph{Advances in Neural Information Processing Systems}, 37:\penalty0 84746--84776, 2024.

\bibitem[Gao et~al.(2025)Gao, Liu, Li, Li, Chen, Zhou, and Metaxas]{gao2025show}
Yunhe Gao, Di~Liu, Zhuowei Li, Yunsheng Li, Dongdong Chen, Mu~Zhou, and Dimitris~N Metaxas.
\newblock Show and segment: Universal medical image segmentation via in-context learning.
\newblock In \emph{Proceedings of the Computer Vision and Pattern Recognition Conference}, pages 20830--20840, 2025.

\bibitem[Hu et~al.(2024)Hu, Shang, Yang, Guo, Peng, and Ma]{hu2024icl}
Jiesi Hu, Yang Shang, Yanwu Yang, Xutao Guo, Hanyang Peng, and Ting Ma.
\newblock Icl-sam: Synergizing in-context learning model and sam in medical image segmentation.
\newblock \emph{Medical Imaging with Deep Learning}, pages 641--656, 2024.

\bibitem[Hu et~al.(2025)Hu, Peng, Yang, Guo, Shang, Shi, Ye, and Ma]{hu2025building}
Jiesi Hu, Hanyang Peng, Yanwu Yang, Xutao Guo, Yang Shang, Pengcheng Shi, Chenfei Ye, and Ting Ma.
\newblock Building 3d in-context learning universal model in neuroimaging.
\newblock \emph{arXiv preprint arXiv:2503.02410}, 2025.

\bibitem[Ilharco et~al.(2022)Ilharco, Ribeiro, Wortsman, Gururangan, Schmidt, Hajishirzi, and Farhadi]{ilharco2022editing}
Gabriel Ilharco, Marco~Tulio Ribeiro, Mitchell Wortsman, Suchin Gururangan, Ludwig Schmidt, Hannaneh Hajishirzi, and Ali Farhadi.
\newblock Editing models with task arithmetic.
\newblock \emph{arXiv preprint arXiv:2212.04089}, 2022.

\bibitem[Isensee et~al.(2021)Isensee, Jaeger, Kohl, Petersen, and Maier-Hein]{isensee2021nnu}
Fabian Isensee, Paul~F Jaeger, Simon~AA Kohl, Jens Petersen, and Klaus~H Maier-Hein.
\newblock nnu-net: a self-configuring method for deep learning-based biomedical image segmentation.
\newblock \emph{Nature methods}, 18\penalty0 (2):\penalty0 203--211, 2021.

\bibitem[Jack~Jr et~al.(2008)Jack~Jr, Bernstein, Fox, Thompson, Alexander, Harvey, Borowski, Britson, L.~Whitwell, Ward, et~al.]{jack2008alzheimer}
Clifford~R Jack~Jr, Matt~A Bernstein, Nick~C Fox, Paul Thompson, Gene Alexander, Danielle Harvey, Bret Borowski, Paula~J Britson, Jennifer L.~Whitwell, Chadwick Ward, et~al.
\newblock The alzheimer's disease neuroimaging initiative (adni): Mri methods.
\newblock \emph{Journal of Magnetic Resonance Imaging: An Official Journal of the International Society for Magnetic Resonance in Medicine}, 27\penalty0 (4):\penalty0 685--691, 2008.

\bibitem[Jin et~al.(2022)Jin, Ren, Preotiuc-Pietro, and Cheng]{jin2022dataless}
Xisen Jin, Xiang Ren, Daniel Preotiuc-Pietro, and Pengxiang Cheng.
\newblock Dataless knowledge fusion by merging weights of language models.
\newblock \emph{arXiv preprint arXiv:2212.09849}, 2022.

\bibitem[Kemker et~al.(2018)Kemker, McClure, Abitino, Hayes, and Kanan]{kemker2018measuring}
Ronald Kemker, Marc McClure, Angelina Abitino, Tyler Hayes, and Christopher Kanan.
\newblock Measuring catastrophic forgetting in neural networks.
\newblock In \emph{Proceedings of the AAAI conference on artificial intelligence}, volume~32, 2018.

\bibitem[Kirillov et~al.(2023)Kirillov, Mintun, Ravi, Mao, Rolland, Gustafson, Xiao, Whitehead, Berg, Lo, et~al.]{kirillov2023segment}
Alexander Kirillov, Eric Mintun, Nikhila Ravi, Hanzi Mao, Chloe Rolland, Laura Gustafson, Tete Xiao, Spencer Whitehead, Alexander~C Berg, Wan-Yen Lo, et~al.
\newblock Segment anything.
\newblock In \emph{Proceedings of the IEEE/CVF international conference on computer vision}, pages 4015--4026, 2023.

\bibitem[Knowles(2006)]{knowles2006parego}
Joshua Knowles.
\newblock Parego: A hybrid algorithm with on-line landscape approximation for expensive multiobjective optimization problems.
\newblock \emph{IEEE transactions on evolutionary computation}, 10\penalty0 (1):\penalty0 50--66, 2006.

\bibitem[Li et~al.(2020)Li, Wang, Wan, Wang, Li, and Kot]{li2020domain}
Haoliang Li, YuFei Wang, Renjie Wan, Shiqi Wang, Tie-Qiang Li, and Alex Kot.
\newblock Domain generalization for medical imaging classification with linear-dependency regularization.
\newblock \emph{Advances in neural information processing systems}, 33:\penalty0 3118--3129, 2020.

\bibitem[Lindauer et~al.(2022)Lindauer, Eggensperger, Feurer, Biedenkapp, Deng, Benjamins, Ruhkopf, Sass, and Hutter]{lindauer2022smac3}
Marius Lindauer, Katharina Eggensperger, Matthias Feurer, Andr{\'e} Biedenkapp, Difan Deng, Carolin Benjamins, Tim Ruhkopf, Ren{\'e} Sass, and Frank Hutter.
\newblock Smac3: A versatile bayesian optimization package for hyperparameter optimization.
\newblock \emph{Journal of Machine Learning Research}, 23\penalty0 (54):\penalty0 1--9, 2022.

\bibitem[Litjens et~al.(2014)Litjens, Toth, Van De~Ven, Hoeks, Kerkstra, Van~Ginneken, Vincent, Guillard, Birbeck, Zhang, et~al.]{litjens2014evaluation}
Geert Litjens, Robert Toth, Wendy Van De~Ven, Caroline Hoeks, Sjoerd Kerkstra, Bram Van~Ginneken, Graham Vincent, Gwenael Guillard, Neil Birbeck, Jindang Zhang, et~al.
\newblock Evaluation of prostate segmentation algorithms for mri: the promise12 challenge.
\newblock \emph{Medical image analysis}, 18\penalty0 (2):\penalty0 359--373, 2014.

\bibitem[Luo et~al.(2024)Luo, Li, Zhang, Liao, and Wang]{luo2024rethinking}
Xiangde Luo, Zihan Li, Shaoting Zhang, Wenjun Liao, and Guotai Wang.
\newblock Rethinking abdominal organ segmentation (raos) in the clinical scenario: A robustness evaluation benchmark with challenging cases.
\newblock In \emph{International Conference on Medical Image Computing and Computer-Assisted Intervention}, pages 531--541. Springer, 2024.

\bibitem[Ma et~al.(2024{\natexlab{a}})Ma, He, Li, Han, You, and Wang]{ma2024segment}
Jun Ma, Yuting He, Feifei Li, Lin Han, Chenyu You, and Bo~Wang.
\newblock Segment anything in medical images.
\newblock \emph{Nature Communications}, 15\penalty0 (1):\penalty0 654, 2024{\natexlab{a}}.

\bibitem[Ma et~al.(2024{\natexlab{b}})Ma, Zhang, Gu, Ge, Mae, Young, Zhu, Yang, Meng, Huang, et~al.]{ma2024unleashing}
Jun Ma, Yao Zhang, Song Gu, Cheng Ge, Shihao Mae, Adamo Young, Cheng Zhu, Xin Yang, Kangkang Meng, Ziyan Huang, et~al.
\newblock Unleashing the strengths of unlabelled data in deep learning-assisted pan-cancer abdominal organ quantification: the flare22 challenge.
\newblock \emph{The Lancet Digital Health}, 6\penalty0 (11):\penalty0 e815--e826, 2024{\natexlab{b}}.

\bibitem[Maron et~al.(2022)Maron, Hekler, Haggenm{\"u}ller, von Kalle, Utikal, M{\"u}ller, Gaiser, Meier, Hobelsberger, Gellrich, et~al.]{maron2022model}
Roman~C Maron, Achim Hekler, Sarah Haggenm{\"u}ller, Christof von Kalle, Jochen~S Utikal, Verena M{\"u}ller, Maria Gaiser, Friedegund Meier, Sarah Hobelsberger, Frank~F Gellrich, et~al.
\newblock Model soups improve performance of dermoscopic skin cancer classifiers.
\newblock \emph{European Journal of Cancer}, 173:\penalty0 307--316, 2022.

\bibitem[Matena and Raffel(2022)]{matena2022merging}
Michael~S Matena and Colin~A Raffel.
\newblock Merging models with fisher-weighted averaging.
\newblock \emph{Advances in Neural Information Processing Systems}, 35:\penalty0 17703--17716, 2022.

\bibitem[Menze et~al.(2014)Menze, Jakab, Bauer, Kalpathy-Cramer, Farahani, Kirby, Burren, Porz, Slotboom, Wiest, et~al.]{menze2014multimodal}
Bjoern~H Menze, Andras Jakab, Stefan Bauer, Jayashree Kalpathy-Cramer, Keyvan Farahani, Justin Kirby, Yuliya Burren, Nicole Porz, Johannes Slotboom, Roland Wiest, et~al.
\newblock The multimodal brain tumor image segmentation benchmark (brats).
\newblock \emph{IEEE transactions on medical imaging}, 34\penalty0 (10):\penalty0 1993--2024, 2014.

\bibitem[Neyshabur et~al.(2020)Neyshabur, Sedghi, and Zhang]{neyshabur2020being}
Behnam Neyshabur, Hanie Sedghi, and Chiyuan Zhang.
\newblock What is being transferred in transfer learning?
\newblock \emph{Advances in neural information processing systems}, 33:\penalty0 512--523, 2020.

\bibitem[Qazi et~al.(2024)Qazi, Almakky, Hashmi, Sanjeev, and Yaqub]{qazi2024dynammo}
Mohammad~Areeb Qazi, Ibrahim Almakky, Anees Ur~Rehman Hashmi, Santosh Sanjeev, and Mohammad Yaqub.
\newblock Dynammo: Dynamic model merging for efficient class incremental learning for medical images.
\newblock In \emph{Annual Conference on Medical Image Understanding and Analysis}, pages 245--257. Springer, 2024.

\bibitem[Rakic et~al.(2024)Rakic, Wong, Ortiz, Cimini, Guttag, and Dalca]{rakic2024tyche}
Marianne Rakic, Hallee~E Wong, Jose Javier~Gonzalez Ortiz, Beth~A Cimini, John~V Guttag, and Adrian~V Dalca.
\newblock Tyche: Stochastic in-context learning for medical image segmentation.
\newblock In \emph{Proceedings of the IEEE/CVF Conference on Computer Vision and Pattern Recognition}, pages 11159--11173, 2024.

\bibitem[Rosenhain et~al.(2018)Rosenhain, Magnuska, Yamoah, Rawashdeh, Kiessling, Gremse, et~al.]{rosenhain2018preclinical}
Stefanie Rosenhain, Zuzanna~A Magnuska, Grace~G Yamoah, Wa’ Rawashdeh, Fabian Kiessling, Felix Gremse, et~al.
\newblock A preclinical micro-computed tomography database including 3d whole body organ segmentations.
\newblock \emph{Scientific data}, 5\penalty0 (1):\penalty0 1--9, 2018.

\bibitem[Sanjeev et~al.(2024)Sanjeev, Zhaksylyk, Almakky, Hashmi, Qazi, and Yaqub]{sanjeev2024fissionfusion}
Santosh Sanjeev, Nuren Zhaksylyk, Ibrahim Almakky, Anees Ur~Rehman Hashmi, Mohammad~Areeb Qazi, and Mohammad Yaqub.
\newblock Fissionfusion: fast geometric generation and hierarchical souping for medical image analysis.
\newblock In \emph{International Conference on Medical Image Computing and Computer-Assisted Intervention}, pages 131--141. Springer, 2024.

\bibitem[Staal et~al.(2004)Staal, Abr{\`a}moff, Niemeijer, Viergever, and Van~Ginneken]{staal2004ridge}
Joes Staal, Michael~D Abr{\`a}moff, Meindert Niemeijer, Max~A Viergever, and Bram Van~Ginneken.
\newblock Ridge-based vessel segmentation in color images of the retina.
\newblock \emph{IEEE transactions on medical imaging}, 23\penalty0 (4):\penalty0 501--509, 2004.

\bibitem[Su and Geiping(2025)]{su2025fine}
Guinan Su and Jonas Geiping.
\newblock Fine, i'll merge it myself: A multi-fidelity framework for automated model merging.
\newblock \emph{arXiv preprint arXiv:2502.04030}, 2025.

\bibitem[Su et~al.(2025)Su, Shen, Yin, Liu, Yang, and Geiping]{su2025gptailor}
Guinan Su, Li~Shen, Lu~Yin, Shiwei Liu, Yanwu Yang, and Jonas Geiping.
\newblock Gptailor: Large language model pruning through layer cutting and stitching.
\newblock \emph{arXiv preprint arXiv:2506.20480}, 2025.

\bibitem[Takaya and Yamamoto(2024)]{takaya2024context}
Eichi Takaya and Shinnosuke Yamamoto.
\newblock In-context learning for medical image segmentation.
\newblock \emph{arXiv preprint arXiv:2412.13299}, 2024.

\bibitem[Utans(1996)]{utans1996weight}
Joachim Utans.
\newblock Weight averaging for neural networks and local resampling schemes.
\newblock In \emph{Proc. AAAI-96 Workshop on Integrating Multiple Learned Models. AAAI Press}, pages 133--138. Citeseer, 1996.

\bibitem[Wang et~al.(2024)Wang, Guo, Ye, Deng, Cheng, Li, Chen, Su, Huang, Shen, et~al.]{wang2024sam}
Haoyu Wang, Sizheng Guo, Jin Ye, Zhongying Deng, Junlong Cheng, Tianbin Li, Jianpin Chen, Yanzhou Su, Ziyan Huang, Yiqing Shen, et~al.
\newblock Sam-med3d: towards general-purpose segmentation models for volumetric medical images.
\newblock In \emph{European Conference on Computer Vision}, pages 51--67. Springer, 2024.

\bibitem[Wang et~al.(2023)Wang, Zhang, Cao, Wang, Shen, and Huang]{wang2023seggpt}
Xinlong Wang, Xiaosong Zhang, Yue Cao, Wen Wang, Chunhua Shen, and Tiejun Huang.
\newblock Seggpt: Towards segmenting everything in context.
\newblock In \emph{Proceedings of the IEEE/CVF International Conference on Computer Vision}, pages 1130--1140, 2023.

\bibitem[White(2016)]{white2016sampling}
Tom White.
\newblock Sampling generative networks.
\newblock \emph{arXiv preprint arXiv:1609.04468}, 2016.

\bibitem[Wu et~al.(2025)Wu, Wang, Hong, Ji, Fu, Xu, Xu, and Jin]{wu2025medical}
Junde Wu, Ziyue Wang, Mingxuan Hong, Wei Ji, Huazhu Fu, Yanwu Xu, Min Xu, and Yueming Jin.
\newblock Medical sam adapter: Adapting segment anything model for medical image segmentation.
\newblock \emph{Medical image analysis}, 102:\penalty0 103547, 2025.

\bibitem[Yadav et~al.(2024)Yadav, Tam, Choshen, Raffel, and Bansal]{yadav2024ties}
Prateek Yadav, Derek Tam, Leshem Choshen, Colin~A Raffel, and Mohit Bansal.
\newblock Ties-merging: Resolving interference when merging models.
\newblock \emph{Advances in Neural Information Processing Systems}, 36, 2024.

\bibitem[Yang et~al.(2023)Yang, Wang, Shen, Liu, Guo, Wang, and Tao]{yang2023adamerging}
Enneng Yang, Zhenyi Wang, Li~Shen, Shiwei Liu, Guibing Guo, Xingwei Wang, and Dacheng Tao.
\newblock Adamerging: Adaptive model merging for multi-task learning.
\newblock \emph{arXiv preprint arXiv:2310.02575}, 2023.

\bibitem[Yang et~al.(2024)Yang, Musio, Ma, Juchler, Paetzold, Al-Maskari, H{\"o}her, Li, Hamamci, Sekuboyina, et~al.]{topcowdata}
Kaiyuan Yang, Fabio Musio, Yihui Ma, Norman Juchler, Johannes~C Paetzold, Rami Al-Maskari, Luciano H{\"o}her, Hongwei~Bran Li, Ibrahim~Ethem Hamamci, Anjany Sekuboyina, et~al.
\newblock Benchmarking the cow with the topcow challenge: Topology-aware anatomical segmentation of the circle of willis for cta and mra.
\newblock \emph{ArXiv}, pages arXiv--2312, 2024.

\bibitem[Ye et~al.(2023)Ye, Cheng, Chen, Deng, Li, Wang, Su, Huang, Chen, Jiang, et~al.]{ye2023sa}
Jin Ye, Junlong Cheng, Jianpin Chen, Zhongying Deng, Tianbin Li, Haoyu Wang, Yanzhou Su, Ziyan Huang, Jilong Chen, Lei Jiang, et~al.
\newblock Sa-med2d-20m dataset: Segment anything in 2d medical imaging with 20 million masks.
\newblock \emph{arXiv preprint arXiv:2311.11969}, 2023.

\bibitem[Yu et~al.(2024)Yu, Yu, Yu, Huang, and Li]{yu2024language}
Le~Yu, Bowen Yu, Haiyang Yu, Fei Huang, and Yongbin Li.
\newblock Language models are super mario: Absorbing abilities from homologous models as a free lunch.
\newblock In \emph{Forty-first International Conference on Machine Learning}, 2024.

\bibitem[Zhang et~al.(2024)Zhang, Wang, Pan, Jiang, Ge, Guo, Jiang, Lu, Zhang, Liu, et~al.]{zhang2024nasalseg}
Yichi Zhang, Jing Wang, Tan Pan, Quanling Jiang, Jingjie Ge, Xin Guo, Chen Jiang, Jie Lu, Jianning Zhang, Xueling Liu, et~al.
\newblock Nasalseg: A dataset for automatic segmentation of nasal cavity and paranasal sinuses from 3d ct images.
\newblock \emph{Scientific Data}, 11\penalty0 (1):\penalty0 1329, 2024.

\bibitem[Zhao et~al.(2025)Zhao, Yang, Li, Jiao, Zhai, Li, Wu, Fu, and Cheng]{zhao2025segmic}
Jianwei Zhao, Fan Yang, Xin Li, Zicheng Jiao, Qiang Zhai, Xiaomeng Li, De~Wu, Huazhu Fu, and Hong Cheng.
\newblock Segmic: A universal model for medical image segmentation through in-context learning.
\newblock \emph{Pattern Recognition}, page 112179, 2025.

\end{thebibliography}
\endgroup

\appendix
\section{Experiments}
\subsection{Data description}
\begin{table}[h]
  \centering
  \setlength{\tabcolsep}{14pt}
  \scriptsize
  \caption{Detailed description of data across 25 tasks.}
\begin{tabular}{ccccccc}
\hline
\hline
idx   & Task Name & dataset & Type  & Organ & Modality & Scans \bigstrut\\
\hline
1     & Brain Tumor & Brats & 3D    & Brain & T1, T2 & 400 \bigstrut[t]\\
2     & Vascular & Topcow & 3D    & Brain & MRA   & 90 \\
3     & Cerebral Cortex & ADNI  & 3D    & Brain & T1    & 400 \\
4     & Hippocampus & ADNI  & 3D    & Brain & T1    & 400 \\
5     & Thalamus & ADNI  & 3D    & Brain & T1    & 400 \\
6     & Lateral Ventricle & ADNI  & 3D    & Brain & T1    & 400 \\
7     & Putamen & ADNI  & 3D    & Brain & T1    & 400 \\
8     & Amygdala & ADNI  & 3D    & Brain & T1    & 400 \\
9     & FLARE22 Liver & FLARE22 & 3D    & Liver & CT    & 50 \\
10    & FLARE22 Kidney\_R & FLARE22 & 3D    & Kidney & CT    & 50 \\
11    & FLARE22 Kidney\_L & FLARE22 & 3D    & Kidney & CT    & 50 \\
12    & Maxillary Sinus & Nasal & 3D    & Nasal & CT    & 120 \\
13    & Nasal Cavity & Nasal & 3D    & Nasal & CT    & 120 \\
14    & Nasal Pharynx & Nasal & 3D    & Nasal & CT    & 120 \\
15    & Prostate & Promise12 & 3D    & Prostate & T2    & 50 \\
16    & Mice-Lung & Mice  & 3D    & Lung  & CT    & 40 \\
17    & Mice-Pancreas & Mice  & 3D    & Pancreas & CT    & 40 \\
18    & Cardiac & MSD-Heart & 3D    & Cardiac & CT    & 18 \\
19    & FLARE22 Spleen & FLARE22 & 3D    & Spleen & CT    & 50 \\
20    & FLARE22 Pancreas & FLARE22 & 3D    & Pancreas & CT    & 50 \\
21    & RAOS Liver & RAOS  & 3D    & Liver & CT    & 262 \\
22    & RAOS Kidney & RAOS  & 3D    & Kidney & CT    & 262 \\
23    & RAOS Stomach & RAOS  & 3D    & Stomach & CT    & 262 \\
24    & Optic Cup & Fundus & 2D    & Fundus & OCT   & 101 \\
25    & Optic Disk & Fundus & 2D    & Fundus & OCT   & 101 \bigstrut[b]\\
\hline
\hline
\end{tabular}%

  \label{tab_data}%
\end{table}%

We provide detailed data descriptions in Table~\ref{tab_data}, covering 25 tasks. For each dataset and task, we split the data into 80\% for evaluation (test set) and 20\% as calibration set. It is important to note that the sizes reported refer to the number of 3D images. Since our study conducts comparisons in a 2D setting for consistency, each 3D image is further divided into hundreds of slices.

\subsection{Additional explanation for model selection}
Since our model merging framework is based on the assumption of same loss basins, models with differing architectures are not considered as candidate models. For example, SAM-Med2D \cite{ye2023sa} employs an adapter-based fine-tuning approach, which introduces architectural modifications that deviate from the base structure. Therefore, we select SAM as the base model and use MedSAM and MedicoSAM, both fine-tuned variants with consistent architectures, as candidate models for merging.

\subsection{SMAC sampling process}
SMAC~\cite{lindauer2022smac3} employs random forests (RF) as surrogate models to guide the configuration sampling process. The workflow operates as follows:

(1) Surrogate Model Training: Random forests are trained to predict model performance based on previously evaluated configurations, including layer-wise merging selections and model combination strategies.

(2) Acquisition-based Sampling: Leveraging the predictions and uncertainty estimates from the surrogate model, SMAC samples new configurations that balance exploitation (favoring configurations with high predicted performance) and exploration (targeting configurations with high predictive uncertainty).

(3) Iterative Refinement: As additional configurations are evaluated, the surrogate model is incrementally updated, progressively enhancing its predictive accuracy and enabling more effective sampling in subsequent iterations.

\subsection{Additional comparison}
In addition to the baselines presented in the main document, we provide supplementary comparison results with the ICL-based model Neuralizer \cite{czolbe2023neuralizer} and the SAM-based model SAM-Med3D \cite{wang2024sam}. The detailed results are reported in Section B-Results.

\begin{table*}[h]
  \centering
  \scriptsize
  \caption{Additional comparison results in terms of Dice coefficient across 25 tasks.}
  \setlength{\tabcolsep}{4pt}
    \begin{tabular}{cccccccccccccccc}
    \hline
    \hline
    \multirow{2}[4]{*}{idx} &       & \multirow{2}[4]{*}{Task Name} &       & ICL &       & SAM &       & \multicolumn{4}{c}{model merging baselines} &       & \multicolumn{3}{c}{MedSAMix-M} \bigstrut\\
\cline{5-5}\cline{7-7}\cline{9-12}\cline{14-16}          &       &       &       & Neuralizer &       & \makecell{SAM-\\ Med3D} &       & SLERP & TA    & TIES  & Linear &       & run1  & run2  & run3 \bigstrut\\
\cline{1-1}\cline{3-3}\cline{5-5}\cline{7-7}\cline{9-12}\cline{14-16}    1     &       & Brain Tumor &       & 6.54  &       & 46.34 &       & 74.14 & 77.89 & 68.56 & 78.12 &       & 74.90 & 76.20 & 75.20 \bigstrut[t]\\
    2     &       & Vascular &       & 8.19  &       & 6.01  &       & 52.91 & 34.73 & 41.47 & 43.79 &       & 62.90 & 53.70 & 62.90 \\
    3     &       & Cerebral Cortex &       & 67.14 &       & 9.15  &       & 55.85 & 52.27 & 43.50 & 54.19 &       & 54.90 & 55.60 & 55.00 \\
    4     &       & Hippocampus &       & 51.53 &       & 30.75 &       & 59.59 & 49.66 & 40.62 & 52.84 &       & 55.00 & 55.90 & 57.50 \\
    5     &       & Thalamus &       & 71.76 &       & 19.11 &       & 53.38 & 67.63 & 44.53 & 66.29 &       & 67.30 & 69.10 & 67.60 \\
    6     &       & Lateral Ventricle &       & 61.80 &       & 30.00 &       & 76.93 & 48.83 & 60.26 & 64.12 &       & 77.60 & 75.10 & 77.60 \\
    7     &       & Putamen &       & 62.08 &       & 14.03 &       & 36.30 & 38.09 & 27.48 & 37.95 &       & 32.60 & 37.40 & 33.60 \\
    8     &       & Amygdala &       & 46.24 &       & 11.02 &       & 46.20 & 49.90 & 36.80 & 51.20 &       & 51.50 & 53.40 & 52.00 \\
    9     &       & FLARE22 Liver &       & 62.64 &       & 70.62 &       & 92.79 & 92.40 & 90.38 & 93.61 &       & 92.20 & 90.80 & 92.90 \\
    10    &       & FLARE22 Kidney\_R &       & 58.97 &       & 84.83 &       & 95.15 & 92.99 & 92.44 & 94.45 &       & 95.40 & 95.20 & 95.10 \\
    11    &       & FLARE22 Kidney\_L &       & 55.80 &       & 88.33 &       & 94.47 & 92.45 & 91.35 & 94.14 &       & 94.50 & 94.20 & 94.20 \\
    12    &       & Maxillary Sinus &       & 74.31 &       & 39.75 &       & 79.01 & 74.05 & 56.36 & 75.91 &       & 80.50 & 80.50 & 79.60 \\
    13    &       & Nasal Cavity &       & 56.45 &       & 15.07 &       & 60.16 & 46.97 & 46.77 & 52.34 &       & 64.70 & 62.80 & 65.30 \\
    14    &       & Nasal Pharynx &       & 76.76 &       & 48.66 &       & 86.57 & 83.02 & 82.73 & 87.08 &       & 89.20 & 87.60 & 88.80 \\
    15    &       & Prostate &       & 71.30 &       & 62.79 &       & 89.16 & 86.83 & 80.48 & 88.36 &       & 91.50 & 91.90 & 90.60 \\
    16    &       & Mice-Lung &       & 66.50 &       & 21.21 &       & 79.26 & 65.87 & 66.98 & 68.91 &       & 78.00 & 72.70 & 75.40 \\
    17    &       & Mice-Pancreas &       & 40.16 &       & 8.40  &       & 87.43 & 85.35 & 83.34 & 86.54 &       & 88.40 & 89.00 & 88.40 \\
    18    &       & Cardiac &       & 41.63 &       & 48.33 &       & 76.95 & 74.77 & 77.20 & 79.41 &       & 84.20 & 84.00 & 83.00 \\
    19    &       & FLARE22 Spleen &       & 43.75 &       & 86.26 &       & 95.03 & 94.06 & 91.98 & 95.04 &       & 94.30 & 93.80 & 94.80 \\
    20    &       & FLARE22 Pancreas &       & 5.76  &       & 22.09 &       & 77.51 & 71.47 & 65.07 & 75.11 &       & 78.00 & 77.80 & 78.40 \\
    21    &       & RAOS Liver &       & 72.95 &       & 60.31 &       & 90.54 & 89.68 & 87.62 & 90.02 &       & 90.00 & 90.10 & 89.80 \\
    22    &       & RAOS Kidney &       & 58.51 &       & 30.42 &       & 71.31 & 73.28 & 52.65 & 71.78 &       & 72.30 & 73.70 & 72.70 \\
    23    &       & RAOS Stomach &       & 27.32 &       & 36.18 &       & 86.65 & 85.74 & 78.10 & 86.39 &       & 86.70 & 87.50 & 86.00 \\
    24    &       & Optic Cup &       & 73.65 &       & -     &       & 74.35 & 83.80 & 64.76 & 80.69 &       & 84.60 & 86.30 & 82.10 \\
    25    &       & Optic Disk &       & 93.74 &       & -     &       & 90.46 & 92.95 & 86.33 & 94.01 &       & 93.70 & 94.10 & 94.70 \bigstrut[b]\\
    \hline
          &       & Avg.  &       & 54.22 &       & 38.68 &       & 75.28 & 72.19 & 66.31 & 74.49 &       & 77.40 & 77.14 & 77.33 \bigstrut\\
    \hline
    \hline
    \end{tabular}%
  \label{supply_main}%
\end{table*}%

\subsection{Descriptions of Existing Model Merging Methods}
\subsubsection{Task Arithmetic} 

Task Arithmetic augments model capabilities by linearly combining task-specific knowledge through vector operations. Given a pre-trained model with weights $\theta_{\text{pre}}$ and a set of fine-tuned task weights ${\theta_{t}^{\text{ft}}}_{t=1}^n$, the task vectors are computed as:

\begin{equation}
\tau_t = \theta_{t}^{\text{ft}} - \theta_{\text{pre}}
\end{equation}

The final merged model weights are derived by applying a weighted sum of these task vectors to the base model:

\begin{equation}
\theta_{\text{Merge}} = \theta_{\text{pre}} + \lambda \sum_{t=1}^n \tau_t
\end{equation}

where $\lambda$ serves as a scaling factor, controlling the extent of task-specific adaptation.

\subsubsection{TIES-Merging} 
TIES-Merging resolves parameter conflicts through a three-step procedure. First, for each task vector $\tau_t$, the top $k\%$ of parameters with the largest magnitudes are selected:

\begin{equation}
\hat{\tau}_t = \text{TopK}(\tau_t, k)
\end{equation}

Next, a consensus sign vector $\hat{\gamma}$ is derived by aggregating the directional trends of parameter changes across tasks:

\begin{equation}
\hat{\gamma} = \text{sgn}\left(\sum_{t=1}^n \hat{\tau}_t\right)
\end{equation}

Finally, the method computes an averaged task vector $\tilde{\tau}$ by considering only those task-specific updates whose signs align with the consensus direction:

\begin{equation}
\tilde{\tau} = \text{Average}({\hat{\tau}_t : \text{sgn}(\hat{\tau}_t) = \hat{\gamma}})
\end{equation}

The merged model weights are then obtained by adding the scaled consensus update to the base model:

\begin{equation}
\theta_{\text{Merge}} = \theta_{\text{pre}} + \lambda \cdot \tilde{\tau}
\end{equation}

\subsubsection{SLERP} 

SLERP (Spherical Linear Interpolation) computes the shortest path on the hypersphere between two sets of model weights, ensuring smooth interpolation in parameter space. Given two models with weights $\theta_1$ and $\theta_2$, the interpolated weights at position $t \in [0,1]$ are calculated as:

\begin{equation}
\text{SLERP}(\theta_1, \theta_2, t) = \frac{\sin((1-t)\omega)}{\sin(\omega)}\theta_1 + \frac{\sin(t\omega)}{\sin(\omega)}\theta_2
\end{equation}

where $\omega = \arccos\left(\frac{\langle\theta_1, \theta_2\rangle}{|\theta_1||\theta_2|}\right)$ represents the angle between the two weight vectors.

\subsubsection{Linear Merging}
Linear Merging performs a simple weighted average of model weights. Given a set of models with weights ${\theta_t}_{t=1}^n$, the merged weights are computed as:

\begin{equation}
\theta_{\text{Linear}} = \sum_{t=1}^n w_t \theta_t
\end{equation}

where the weights satisfy $\sum_{t=1}^n w_t = 1$ and $w_t \geq 0$, ensuring a convex combination.


\begin{table*}[h]
  \centering
  \setlength{\tabcolsep}{6pt}
  \small
  \caption{Architecture of our merged MedSAMix-M}
    \begin{tabular}{cccccccccc}
    \hline
    \hline
    \multirow{2}[4]{*}{Layer} &       & \multirow{2}[4]{*}{Method} &       & \multicolumn{2}{c}{TIES/TA} &       & \multicolumn{3}{c}{Linear/SLERP} \bigstrut\\
\cline{5-6}\cline{8-10}          &       &       &       & Retain ratio & Scaling &       & SAM   & MedicoSAM & MedSAM \bigstrut\\
\cline{1-1}\cline{3-3}\cline{5-6}\cline{8-10}    Patch Embedding &       & TIES  &       & 0.50  & 0.50  &       & -     & -     & - \bigstrut[t]\\
    \hline
    Pos Embedding &       & SLERP &       & -     & -     &       & -     & 0.59  & 0.55 \\
    \hline
    Encoder Transformer-0 &       & \multirow{2}[0]{*}{TIES} &       & \multirow{2}[0]{*}{0.59} & \multirow{2}[0]{*}{0.07} &       & \multirow{2}[0]{*}{-} & \multirow{2}[0]{*}{-} & \multirow{2}[0]{*}{-} \\
    Encoder Transformer-1 &       &       &       &       &       &       &       &       &  \\
    \hline
    Encoder Transformer-2 &       & \multirow{2}[0]{*}{TA} &       & \multirow{2}[0]{*}{-} & \multirow{2}[0]{*}{0.30} &       & \multirow{2}[0]{*}{-} & \multirow{2}[0]{*}{-} & \multirow{2}[0]{*}{-} \\
    Encoder Transformer-3 &       &       &       &       &       &       &       &       &  \\
    \hline
    Encoder Transformer-4 &       & \multirow{2}[0]{*}{Linear} &       & \multirow{2}[0]{*}{-} & \multirow{2}[0]{*}{-} &       & \multirow{2}[0]{*}{0.60} & \multirow{2}[0]{*}{0.59} & \multirow{2}[0]{*}{0.12} \\
    Encoder Transformer-5 &       &       &       &       &       &       &       &       &  \\
    \hline
    Encoder Transformer-6 &       & \multirow{2}[0]{*}{TA} &       & \multirow{2}[0]{*}{-} & \multirow{2}[0]{*}{0.06} &       & \multirow{2}[0]{*}{-} & \multirow{2}[0]{*}{-} & \multirow{2}[0]{*}{-} \\
    Encoder Transformer-7 &       &       &       &       &       &       &       &       &  \\
    \hline
    Encoder Transformer-8 &       & \multirow{2}[0]{*}{SLERP} &       & \multirow{2}[0]{*}{-} & \multirow{2}[0]{*}{-} &       & \multirow{2}[0]{*}{0.93} & \multirow{2}[0]{*}{-} & \multirow{2}[0]{*}{0.90} \\
    Encoder Transformer-9 &       &       &       &       &       &       &       &       &  \\
    \hline
    Encoder Transformer-10 &       & \multirow{2}[0]{*}{TA} &       & \multirow{2}[0]{*}{-} & \multirow{2}[0]{*}{0.06} &       & \multirow{2}[0]{*}{-} & \multirow{2}[0]{*}{-} & \multirow{2}[0]{*}{-} \\
    Encoder Transformer-11 &       &       &       &       &       &       &       &       &  \\
    \hline
    Prompt Encoder-0 &       & SLERP &       & -     & -     &       & 0.95  & -     & 0.75 \\
    \hline
    Prompt Encoder-1 &       & TA    &       & -     & 0.23  &       & -     & -     & - \\
    \hline
    Prompt Encoder-2 &       & SLERP &       & -     & -     &       & 0.77  & -     & 0.67 \\
    \hline
    Prompt Encoder-3 &       & Linear &       & -     & -     &       & 0.95  & 0.54  & 0.57 \\
    \hline
    Neck  &       & TIES  &       & 0.50  & 0.50  &       & -     & -     & - \\
    \hline
    Mask decoder layer-0 &       & Linear &       & -     & -     &       & 0.44  & 0.30  & 0.01 \\
    \hline
    Mask decoder layer-1 &       & Linear &       & -     & -     &       & 0.40  & 0.50  & 0.50 \\
    \hline
    Mask decoder layer-2 &       & Linear &       & -     & -     &       & 0.50  & 0.50  & 0.50 \\
    \hline
    Mask decoder layer-3 &       & SLERP &       & -     & -     &       & 0.50  & 0.50  & - \\
    \hline
    Mask decoder layer-4 &       & TIES  &       & 0.18  & 0.55  &       & -     & -     & - \\
    \hline
    Mask decoder layer-5 &       & SLERP &       & -     & -     &       & 0.50  & -     & 0.50 \bigstrut[b]\\
    \hline
    \hline
    \end{tabular}%
  \label{res_layer}%
\end{table*}%
\subsection{Hyperparameters for model merging candidate methods}
For Task Arithmetic and TIES-Merging, we set the scaling factor controlling the magnitude of task-specific updates within the range $[0.0, 1.0]$. In Linear Merging, the scaling factors are uniformly assigned as $1/n$, where $n$ denotes the number of models being merged. Additionally, for TIES-Merging, we specify a ratio to retain parameters with the largest magnitudes, choosing from $[0.0, 1.0]$.

\section{Results}
\subsection{Additional results}

We provide additional comparison results in Table~\ref{supply_main}, including the ICL-based model Neuralizer, the SAM-based model SAM-Med3D, model merging baselines, and multiple runs of our proposed MedSAMix. We can see that (1) SAM-Med3D underperforms SAM-Med2D across the 25 tasks. This is primarily because SAM-Med2D is fine-tuned on a broader range of datasets, enhancing its generalization capability. In contrast, SAM-Med3D is limited by the scarcity of 3D training data, which allows it to excel in specific tasks such as kidney segmentation but leads to poor generalization in other domains. (2) Baseline model merging methods exhibit inconsistent performance across tasks, indicating their sensitivity to task-specific variations. This inconsistency underscores the limitations of traditional merging approaches that apply uniform strategies across all layers, thereby highlighting the advantage of our MedSAMix, which adopts a flexible, task-adaptive merging mechanism. (3)  Since our approach selects configurations based on Pareto-optimal solutions, we further compare the top three solutions (run-1, run-2, run-3). As shown in Table~\ref{supply_main}, the performance differences among these runs are minimal. MedSAMix consistently achieves stable performance across different optimization runs. Although it is challenging to guarantee the absolute optimal solution in each run, the results remain robust as long as the selected calibration tasks are representative.

\subsection{Merged architecture}
We present the merged architecture of MedSAMix-M in Table~\ref{res_layer}. The results show that different layers adopt different merging strategies, which is a key factor contributing to the superiority of our method. This hierarchical combination allows the model to merge in a direction that is more advantageous for overall performance, enabling a more adaptive and effective integration compared to uniform merging approaches.
\end{document}